\documentclass{article}
\usepackage{spconf,amsmath,graphicx}
\usepackage{amsfonts}
\usepackage[table]{xcolor}
\usepackage{multirow}
\usepackage{tabularx}
\usepackage{url}
\usepackage {caption}
\usepackage{longtable}
\usepackage{array} 
\usepackage{supertabular,booktabs}
\usepackage{boldline,tabu}
\usepackage{color}
\usepackage{arydshln}
\newcommand{\ie}{{\it i.e.}}

\newcolumntype{A}{>{\raggedright\arraybackslash}X[1]}
\newcolumntype{B}{>{\centering\arraybackslash}X[2]}
\newcommand{\eg}{{\it e.g.}}

\newcommand{\Fref}[1]{Fig.~\ref{#1}}

\title{PerFace: Metric Learning in Perceptual Facial Similarity for Enhanced Face Anonymization}
\name{Haruka Kumagai\thanks{Thanks to XYZ agency for funding.}}
\name{Haruka Kumagai$^1$, Leslie Wöhler$^1$, Satoshi Ikehata$^{2,1}$, Kiyoharu Aizawa$^1$}
\address{The University of Tokyo$^1$
NII$^2$}

\begin{document}
\ninept

\maketitle
\begin{abstract}

In response to rising societal awareness of privacy concerns, face anonymization techniques have advanced, including the emergence of face-swapping methods that replace one identity with another. Achieving a balance between anonymity and naturalness in face swapping requires careful selection of identities: overly similar faces compromise anonymity, while dissimilar ones reduce naturalness. Existing models, however, focus on binary identity classification “the same person or not”, making it difficult to measure nuanced similarities such as “completely different” versus “highly similar but different.” This paper proposes a human-perception-based face similarity metric, creating a dataset of 6,400 triplet annotations and metric learning to predict the similarity. Experimental results demonstrate significant improvements in both face similarity prediction and attribute-based face classification tasks over existing methods.
Our dataset is available at \url{https://github.com/kumanotanin/PerFace}.

\end{abstract}
%

\renewcommand{\thefootnote}{\fnsymbol{footnote}}
\footnote[0]{This research was partially supported by JSPS 25H01164.}
\renewcommand{\thefootnote}{\arabic{footnote}}

\section{Introduction}
\label{sec:intro}
Face anonymization is a technique used to protect individual privacy in facial images while maintaining the usability of the underlying information. Among the various approaches such as occlusion, blurring and face swapping—replacing the target's facial identity with the source's while preserving target's identity-irrelevant attributes (e.g., pose, expression, or background)~\cite{deepfakes,chen2020simswap,kim2022diffface,leslie2022faceswap,nirkin2019fsgan,li2019faceshifter,zhu2021one,gao2021information,wang2021hififace,Zhao_2023_CVPR,Shiohara_2023_ICCV}—is considered an effective anonymization method. Unlike traditional methods such as occlusion or blurring, which compromise image quality and utility while effectively concealing the identity of the original person, face swapping ensures realism and coherence by preserving the facial structure in images. 

\begin{figure}[t]
  \centering
  \centerline{\includegraphics[width=8.5cm]{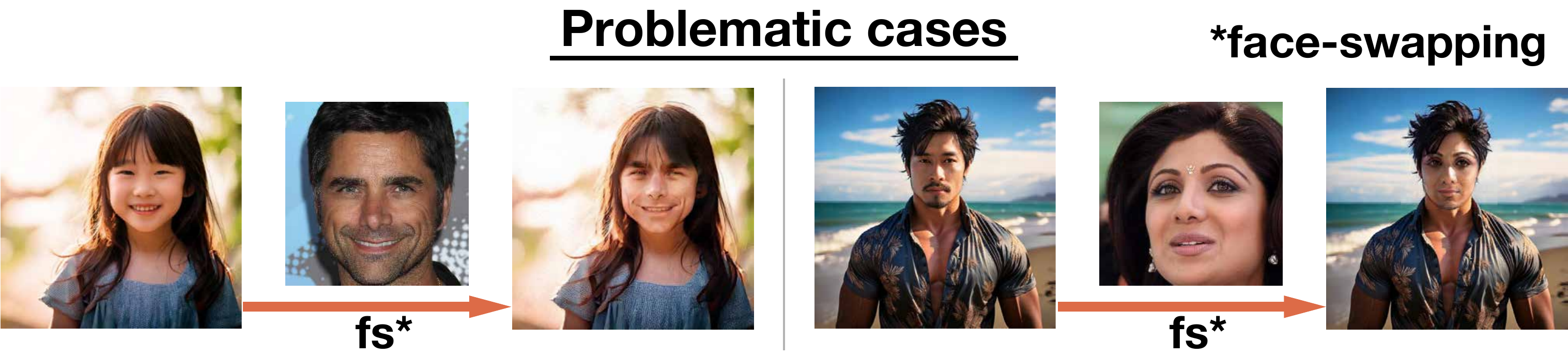}}
\caption{Example of unnatural face swapping. Swapping a young child's face with an adult male's can produce unrealistic images with deep wrinkles and age-inappropriate contours. Similarly, gender-based facial differences can also lead to unnatural results.}
\label{fig:problem}
\end{figure}

When employing face swapping for anonymization, it is crucial that the source and the target share key attributes such as gender and age~\cite{leslie2022faceswap} to avoid unnatural outcomes (See~\Fref{fig:problem}). To address this, a recent study~\cite{ciftci2023my} leveraged distances between embedded features from pretrained face recognition models (e.g., ArcFace~\cite{deng2019arcface}). Their algorithm identifies the most distant face in latent space that nonetheless retains similar attributes (e.g., age, gender), thereby ensuring both anonymity and a natural appearance.

However, leveraging pretrained face recognition models~\cite{sun2014deep,schroff2015facenet,parkhi2015deep,baltruvsaitis2016openface,liu2017sphereface,wang2018cosface,deng2019arcface,Meng_2021_CVPR,Qiu_2021_ICCV,Kim_2022_CVPR,Shiohara_2023_ICCV,alansari2023ghostfacenets} for facial anonymization poses two major challenges. First, these models were optimized via metric learning to cluster images of the same identity and separate those of different identities—thereby considering all different identities as dissimilar, even when they appear perceptually similar. In contrast, face swapping for anonymization necessitates an accurate assessment of similarity between different identities to guarantee that the swapped face is perceptually distinct from the original. Second, existing models were trained exclusively on genuine face images. In face anonymization, it is imperative to evaluate the perceptual distance between the original image and its face-swapped counterpart, rather than merely measuring the distance between the target and source images. This is essential because a face's overall impression is influenced not only by its intrinsic parts (\eg, eyes, noses) but also by factors such as facial contours and hairstyle. Moreover, face-swapped images often contain artifacts and attribute inconsistencies, resulting in a significant domain gap from natural images.

\begin{figure}[t]
  \centering
  \centerline{\includegraphics[width=8.5cm]{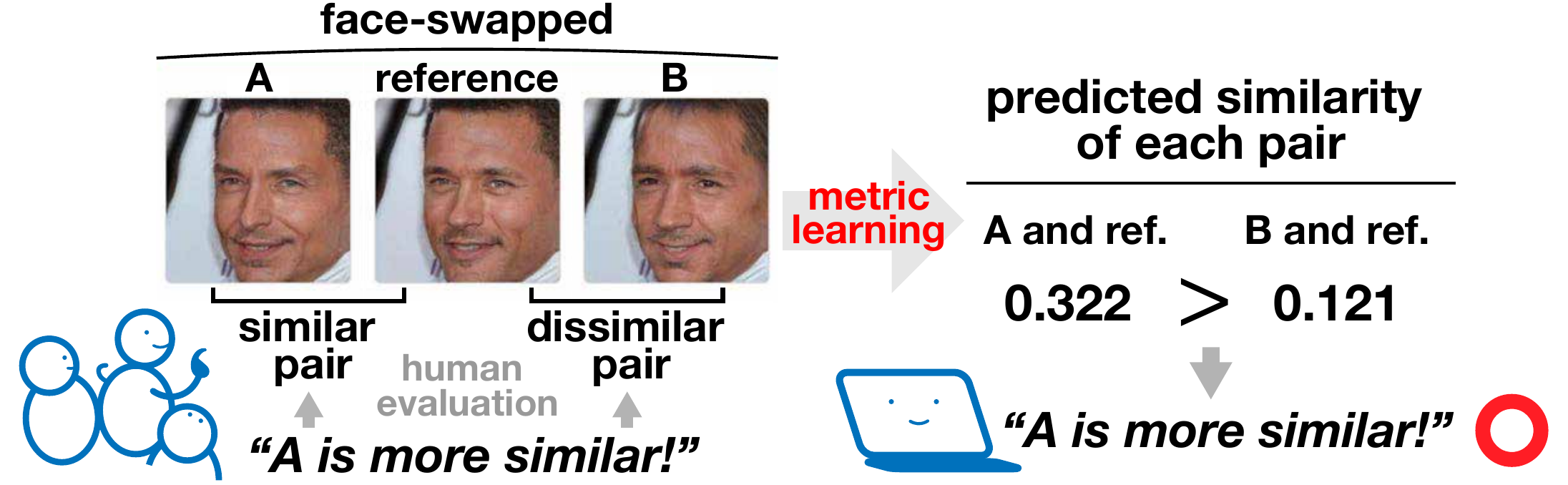}}
\caption{We developed a model that predicts perceptual facial similarity using metric learning, based on our SimCelebA dataset.}
\label{fig:teaser}
\end{figure}

In this work, we present {\it PerFace}, a feature extractor tailored for evaluating facial similarity. Unlike traditional face recognition models that emphasize identity matching, PerFace is trained through metric learning on human-annotated similarity assessments specifically derived from {\it face‐swapped images}. To address the challenge of directly quantifying similarity, we introduced a pairwise comparison task in which annotators select the face‐swapped image that most closely resembles a reference face-swapped image (Fig. \ref{fig:teaser}). Building on PerFace, we propose a comprehensive face anonymization framework that leverages these refined features. Our framework outperforms conventional pretrained models (e.g., ~\cite{deng2019arcface}) on facial similarity assessment and more effectively selects images that are perceptually dissimilar to the original while preserving key attributes such as age and gender. Extensive validation using our human-annotated SimCelebA dataset underscores the effectiveness and specificity of our contributions to face anonymization.

\section{RELATED WORKS}
\label{sec:related-works}

\subsection{Face Recognition Models}
In recent years, machine learning models have come to dominate face recognition, a technology used in applications from smartphone authentication to law enforcement. Typically, these systems identify the face in a database that matches the identity of a given test image. To achieve this, feature extractors are trained to compute distances between feature vectors—proxies for facial similarity—using losses such as contrastive loss~\cite{sun2014deep,sun2015deeply,sun2015deepid3,yi2014learning}, triplet loss~\cite{schroff2015facenet,parkhi2015deep,ding2015robust}, and angular margin-based losses~\cite{deng2019arcface,wang2018cosface,liu2017sphereface}. However, because these similarity measures are optimized for verification and identification—tasks focused on confirming whether two faces share the same identity—they fall short when it comes to effectively quantifying the degree of difference between faces of different individuals.

\subsection{Face Anonymization via Face Swapping}
Face swapping, in the context of facial anonymization, replaces only identity-specific facial features while preserving non-identifying attributes such as expression, pose, and lighting \cite{deepfakes,chen2020simswap,kim2022diffface,leslie2022faceswap,nirkin2019fsgan,li2019faceshifter,zhu2021one,gao2021information,wang2021hififace,Zhao_2023_CVPR,Shiohara_2023_ICCV}. In contrast to conventional anonymization methods—such as occlusion or blurring—that obscure the entire face and often degrade image quality, face swapping modifies only the features critical for identity recognition. In practical applications, this process typically involves replacing a target face with one drawn from a set of source faces that have sufficiently different identities. Although state-of-the-art face recognition systems can detect and quantify differences between target and source faces through deep feature embeddings~\cite{cao2024face}, human observers may still perceive a high degree of similarity or even believe the target and face-swapped images belong to the same individual, even when the model has assigned them different identities.

\section{Method}
\label{sec:method}
The primary goal of this work is to develop an effective facial feature extractor that accurately measures facial similarity, facilitating the selection of appropriate source and target faces for face-swapping. Building on this, we propose a facial anonymization framework that replaces a target face with a source face that shares key attributes (\eg, gender, age) while ensuring that the original and face-swapped images remain as perceptually distinct as possible. Traditional facial anonymization methods (\eg,~\cite{ciftci2023my}) typically rely on pretrained face recognition models (\eg, ArcFace~\cite{deng2019arcface}). However, these models are not optimized for evaluating similarity between images of entirely different identities or for assessing similarity distances in face-swapped images. To address these limitations, we conduct human assessments of similarity in face-swapped images and construct a dataset, {\it SimCelebA}, to train a facial feature extractor, {\it PerFace}, using metric learning.

\subsection{SimCelebA Dataset}
\label{sec-dataset}
Firstly, we conducted a human assessment to create a training dataset containing {\it face-swapped} images annotated with perceptual facial similarity scores. Since assessing the absolute similarity between two face-swapped images is inherently challenging, we adopted a triplet-based approach. Each triplet comprised three images: a reference image (C) and two comparison images (A and B). Participants were tasked with determining which of the two, A or B, bore a closer resemblance to C.

To prevent similarity judgments from being overly influenced by key facial attributes of the target (\eg, hair style, contour) or artifacts caused by face-swapping, we kept the target face constant and selected three distinct source images for swapping. This ensured that all images presented to the participants were face-swapped, eliminating potential biases arising from the presence of natural images.

We utilized the CelebAMask-HQ dataset~\cite{CelebAMask-HQ}, a large-scale collection of 30,000 high-resolution face images. From this dataset, we manually selected 80 target images (40 male and 40 female) and selected 240 source images (120 male and 120 female), forming 80 triplets. To minimize bias, images with obstructions such as glasses were excluded from the dataset.
Using these chosen target and source images, we applied the SimSwap method~\cite{chen2020simswap} to generate face-swapped images. This process resulted in a dataset of 6,400 samples.

We recruited 18 participants, ensuring that each triplet was annotated by at least three individuals to capture human-perceived facial similarity. To enhance the reliability of the annotations, we embedded multiple dummy samples within the triplets. In these dummy samples, two images depicted the same individual, ensuring that a careful examination would unambiguously yield the correct answer. Consequently, only annotations from participants who answered all of these dummy samples correctly were considered valid. As a result, each sample ultimately received three high-quality annotations. 
This dataset was divided into training, validation, and test sets for use.

\subsection{PerFace: facial similarity extractor trained on SimCelebA}
\label{subsec-method-finetuning}
Our goal is for the model to acquire the ability to evaluate facial similarity in a human-like manner, going beyond mere identity matching. To achieve this, we fine-tuned ArcFace~\cite{deng2019arcface} on our SimCelebA dataset, leveraging its strong discriminative power in face recognition tasks to better capture human-perceived facial similarity in face-swapped images. Given an annotated triplet (\ie, A, B, and C) of face-swapped images, we propose using the following triplet loss to train the model:
\begin{eqnarray} \label{eq:loss}
\mathcal{L} = \max \left( 0, \left( \cos(x_i, x_i^-) - \cos(x_i, x_i^+) \right) + m \right).\label{subsec-loss}
\end{eqnarray}
Here, \(x_i \in \mathbb{R}^d\) represents the embedded feature of the reference image (\ie, C) in the \(i\)-th sample. \(x_i^+ \in \mathbb{R}^d\) is the embedded feature of the image chosen as more similar to the reference image \(x_i\) by the majority of annotators (\ie., A or B). Similarly, \(x_i^- \in \mathbb{R}^d\) is the feature vector of the image chosen by the minority (\ie, A or B). The feature vector dimension \(d\) is set to 512, following prior studies~\cite{wen2016discriminative,deng2019arcface,liu2017sphereface,wang2018cosface}. \( \cos(\cdot, \cdot) \) denotes cosine similarity, and \(m\) represents the margin. This triplet loss is minimized to ensure that the distance between similar pairs chosen by annotators decreases, while the distance between non-selected pairs increases. As a result, the model is fine-tuned such that the distances between embedded features of similar faces, as perceived by humans, become smaller.
To facilitate face similarity comparisons, our dataset is designed to select relatively similar faces, as described in Sec.\ref{sec-dataset}. Accordingly, the loss function is also designed to focus on relative distances.
\subsection{Face Anonymization with PerFace}
\label{sec-method-attribute}
In most existing work (e.g., ArcFace~\cite{deng2019arcface}), when employing face swapping for anonymization, only the similarity between the source and target faces is evaluated. In contrast, we aim to assess the similarity between the target face and the {\it face-swapped} face using our PerFace feature extractor. However, simply comparing all pairs of source and target faces is prohibitively resource demanding, so we first align facial attributes before and after swapping and then compare faces only within each group.

\begin{figure}[t]
  \centering
  \centerline{\includegraphics[width=8.5cm]{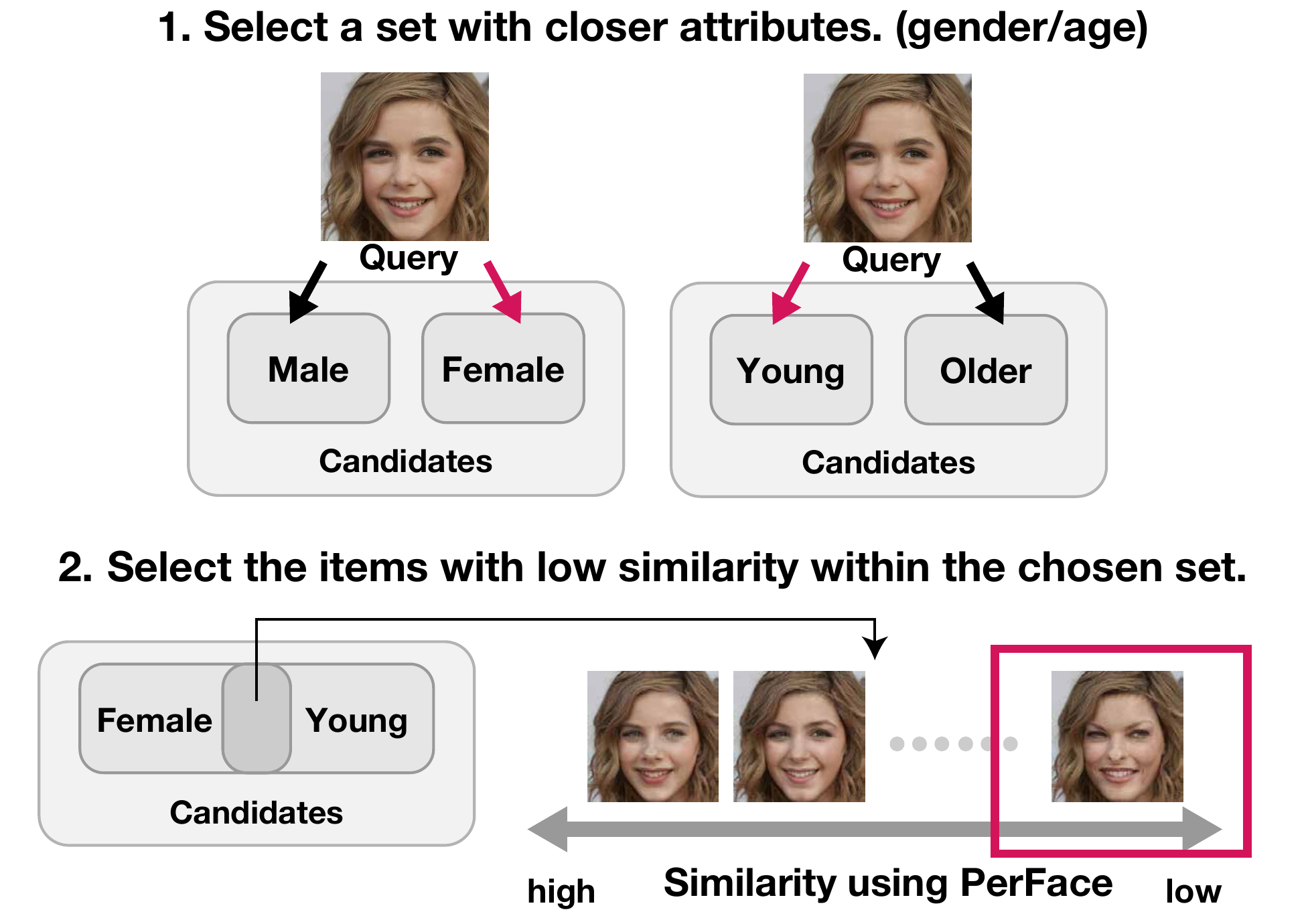}}
\caption{Overview of the method for selecting a suitable source for face swapping with a query image. }
\label{fig:attr-p2}
\end{figure}

The overview of the selection method, which consists of two steps, is shown in Fig.~\ref{fig:attr-p2}.
Considering practical applications, a pre-defined set of face-swap candidates may sometimes be prepared. 
Therefore, this study assumes that the face-swap candidates have been annotated with attributes in advance. Conversely, it is assumed that the target attributes are unknown. Since it is impossible to predict which face the user wishes to anonymize, pre-annotating the target is impractical.

\label{subsec-method-group}

\noindent\textbf{STEP 1: Group Selection.}  
To ensure that the face-swapping process remains natural, suitable face-swapping candidates are identified through facial similarity evaluation.  
Assume there is a face image that the user wants to anonymize, referred to as the query image. The attributes of the query image are determined.  
Within the set of face-swapping candidates, images are grouped based on attributes. Following previous studies~\cite{kim2022smooth}, age and gender are considered as attributes. We created 4 attribute groups (male, female, young, and older) and their intersection sets considering both gender and age (e.g.,young$\cap$male), resulting in a total of 8 groups.
For each attribute group, the similarity with the query image is calculated, and the group with the highest similarity is selected as the face-swapping candidates.

\noindent\textbf{STEP 2: Anonymization.}  
\label{subsec-method-annonimize}  
Next, the anonymization process is performed.  
As long as the selected group from STEP1 is used, the face-swapped image is expected to maintain a certain degree of naturalness. The goal is to achieve anonymization while maintaining this naturalness.  
Using the proposed method, our model trained via metric learning is employed to estimate the similarity between the images of the query and the face-swapping candidates within the group. At this stage, an image with low similarity in the group is chosen as the source for face swapping.  
By choosing less similar face from the same attribute group,
face anonymization can be achieved while ensuring naturalness.

\section{Experiments and Discussion}
\label{sec:experiments}

\subsection{Training details}
We adopted ArcFace \cite{deng2019arcface} as the pretrained model, which we trained using the MS1MV3 \cite{deng2019lightweight} dataset.
For the feature extractor, we employed a ResNet50\cite{he2016deep}.
We considered two cases for training: (1) using all triplet data in the training set ($D1$), and (2) using only triplet data with consistent annotations ($D2$). 




\noindent\textbf{Common Settings for Training Data D1 and D2.}  
The batch size was 32, momentum was set to $0.9$, weight decay to $5$e-$4$, and the learning rate to $0.01$. SGD was used as the optimizer, and the loss margin in Equation \eqref{eq:loss} was set to 0.1.  

\subsection{Evaluation method}
\label{subsec-eval-method}
Using the test triplet data and the annotation results, similar pairs and dissimilar pairs were created.  Similar pairs consisted of the reference image and the image selected as more similar.  Dissimilar pairs consisted of the reference image and the image not selected as more similar.  If the similarity score for the similar pair was higher than that for the dissimilar pair, the model prediction was considered correct for that sample.  
Only samples with consistent annotations across all three responses were adopted as evaluation data.

\subsection{Comparison with other methods}
\begin{figure}[t]
  \centering
  \centerline{\includegraphics[width=8.5cm]{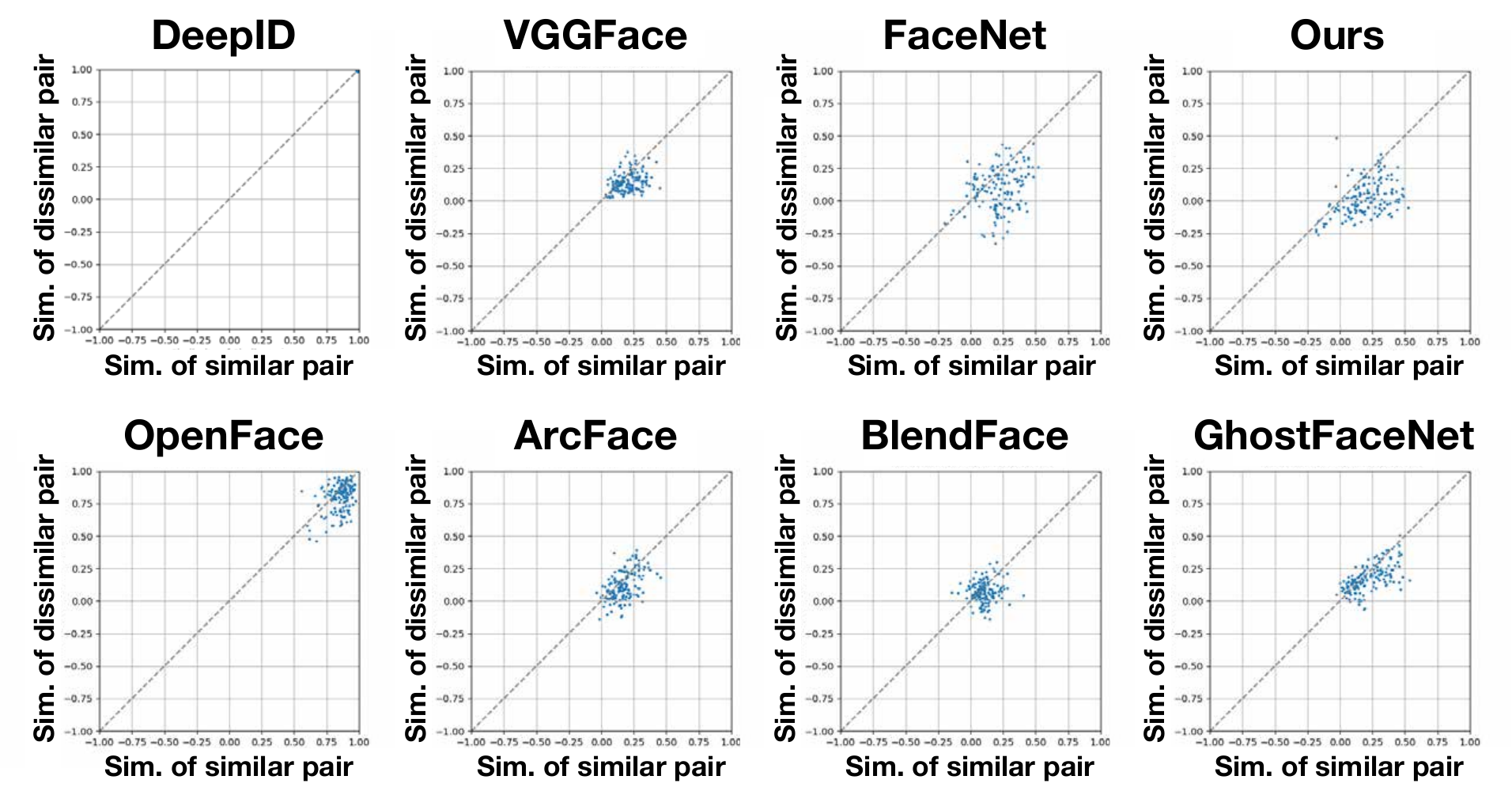}}
\caption{Scatter plot of predicted similarity scores for similar and dissimilar pairs by major face recognition models. The X-axis represents the predicted similarity between similar pairs, while the Y-axis represents the predicted similarity between dissimilar pairs.  
The gray dashed line represents the graph of $Y = X$. Points below $Y < X$ indicate that the pair judged as more similar has been assigned a higher similarity score by the model. }
\label{fig:benchmark}
\end{figure}

Here, we compare our method with existing approaches capable of measuring the distance between faces.
We used $D2$ to ensure the highest annotation quality.
As for BlendFace \cite{Shiohara_2023_ICCV}, we used the official code and weights. Other similarity evaluations were conducted through the DeepFace framework \cite{serengil2020lightface}, with RetinaFace \cite{deng2020retinaface} as the detector and cosine similarity as the metric.


Fig.~\ref{fig:benchmark} presents scatter plots of similarity scores for similar and dissimilar pairs. 
The proportion of points below $Y < X$ corresponds to the accuracy values shown in Table~\ref{tab:benchmark}.  
Our proposed method significantly outperformed existing methods in terms of accuracy.
Models such as ArcFace \cite{deng2019arcface}, VGG Face \cite{parkhi2015deep}, GhostFaceNets \cite{alansari2023ghostfacenets}, and BlendFace\cite{Shiohara_2023_ICCV} predicted similarity scores for both similar and dissimilar pairs around 0.00 to 0.50.  
In contrast, OpenFace \cite{baltruvsaitis2016openface} and DeepID \cite{sun2014deep} predicted high similarity scores for both similar and dissimilar pairs. For the triplet samples used in this evaluation, all three face images inherited the hairstyle and skin tone of target, making it difficult for these models to understand differences in facial parts. 
Specific examples are presented in Table\ref{tab:example-compare}.
Comparison methods make incorrect predictions, often assigning higher similarity scores to the dissimilar pair or consistently predicting nearly identical scores.


\begin{table}[t]

\centering
\caption{Similarity evaluation results in comparison with previous methods.}
    \begin{tabular}{@{}ll@{}}
    \toprule
        \textbf{Method} & \textbf{Acc} \\ \midrule
        DeepID\cite{sun2014deep} & 0.604  \\ 
        VGG Face\cite{parkhi2015deep} & 0.750 \\ 
        FaceNet\cite{schroff2015facenet} & 0.715 \\ 
        OpenFace\cite{baltruvsaitis2016openface} & 0.660 \\ 
        ArcFace\cite{deng2019arcface} & 0.701 \\ 
        BlendFace\cite{Shiohara_2023_ICCV} & 0.576 \\ 
        GhostFaceNets\cite{alansari2023ghostfacenets} & 0.604 \\ \hline \hline
        \textbf{Ours} & \textbf{0.917} \\ \bottomrule
    \end{tabular}
\label{tab:benchmark}
\end{table}

\begin{table}[t]
\centering
\caption{Examples of evaluating similar and dissimilar pairs in triplet samples using our method compared to others. For each triplet, the central image and the left-side image form a similar pair, while the central image and the right-side image form a dissimilar pair.}
\label{tab:example-compare}
\setlength{\tabcolsep}{0pt} 
\begin{tabularx}{\columnwidth}{>{\raggedright\arraybackslash}m{2.75cm}|>{\centering\arraybackslash}X;{2pt/2pt}>{\centering\arraybackslash}X|>{\centering\arraybackslash}X;{2pt/2pt}>{\centering\arraybackslash}X}
\hline
 & similar & dissimilar & similar & dissimilar \\

 \textbf{Method} & \multicolumn{2}{c|}{
   \begin{minipage}{0.33\columnwidth}
   \centering
   \includegraphics[width=\linewidth, trim=0 0 0 0, clip]{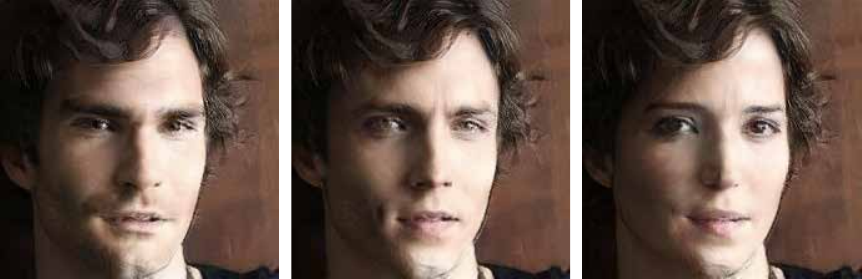}
   \end{minipage}
 } & \multicolumn{2}{c}{
   \begin{minipage}{0.33\columnwidth}
   \centering
   \includegraphics[width=\linewidth,  trim=0 0 0 0, clip]{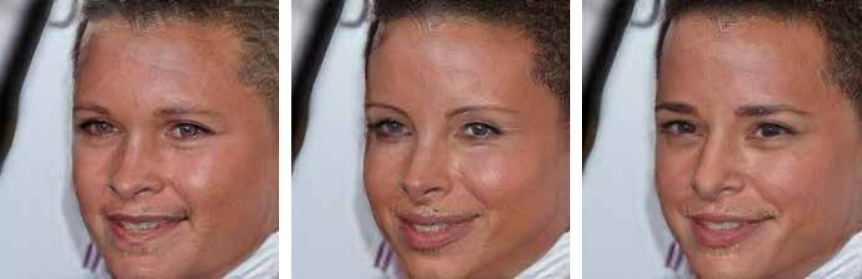}
   \end{minipage}
 } \\
\hline

DeepID\cite{sun2014deep}  & 0.993& 0.997& 0.998& 0.998\\
VGG Face\cite{parkhi2015deep} &0.252 & 0.349& 0.198&0.376 \\
FaceNet\cite{schroff2015facenet} & 0.207&0.261 &0.405 &0.339 \\
OpenFace\cite{baltruvsaitis2016openface} & 0.905& 0.948& 0.864&0.823 \\
ArcFace\cite{deng2019arcface} & 0.287& 0.273& 0.288&0.314 \\
BlendFace\cite{Shiohara_2023_ICCV}& 0.105&0.143& 0.144&0.051 \\
GhostFaceNets\cite{alansari2023ghostfacenets}& 0.333& 0.364& 0.171&0.257 \\
\hline\hline
\textbf{Ours} & \textbf{0.339}& \textbf{-0.006}& \textbf{0.183}&\textbf{0.055} \\
\hline
\end{tabularx}

\end{table}

\subsection{Analyzing the effects on source/target knowledge}
\label{subsec-eval}
\begin{table}[t]
\caption{Mean±SD of facial similarity prediction accuracy.}
\setlength\tabcolsep{2pt} 
\centering
    \begin{tabular}{@{}llll@{}}
    \toprule
        \textbf{Dataset} & \textbf{Acc in [i]} & \textbf{Acc in [ii]} & \textbf{Acc in [iii]} \\ \midrule
        - (Original)     & 0.690 & 0.600   & 0.644 \\  
        D1     & 0.753 $\pm$0.025 & 0.795 $\pm$0.006 & 0.770 $\pm$0.014 \\
        D2     & 0.862 $\pm$0.031 & 0.922 $\pm$0.007 & 0.864 $\pm$0.018 \\ \bottomrule
    \end{tabular}
\label{tab:acc-10}
\end{table}
\begin{figure}[t]
  \centering
  \centerline{\includegraphics[width=8.5cm]{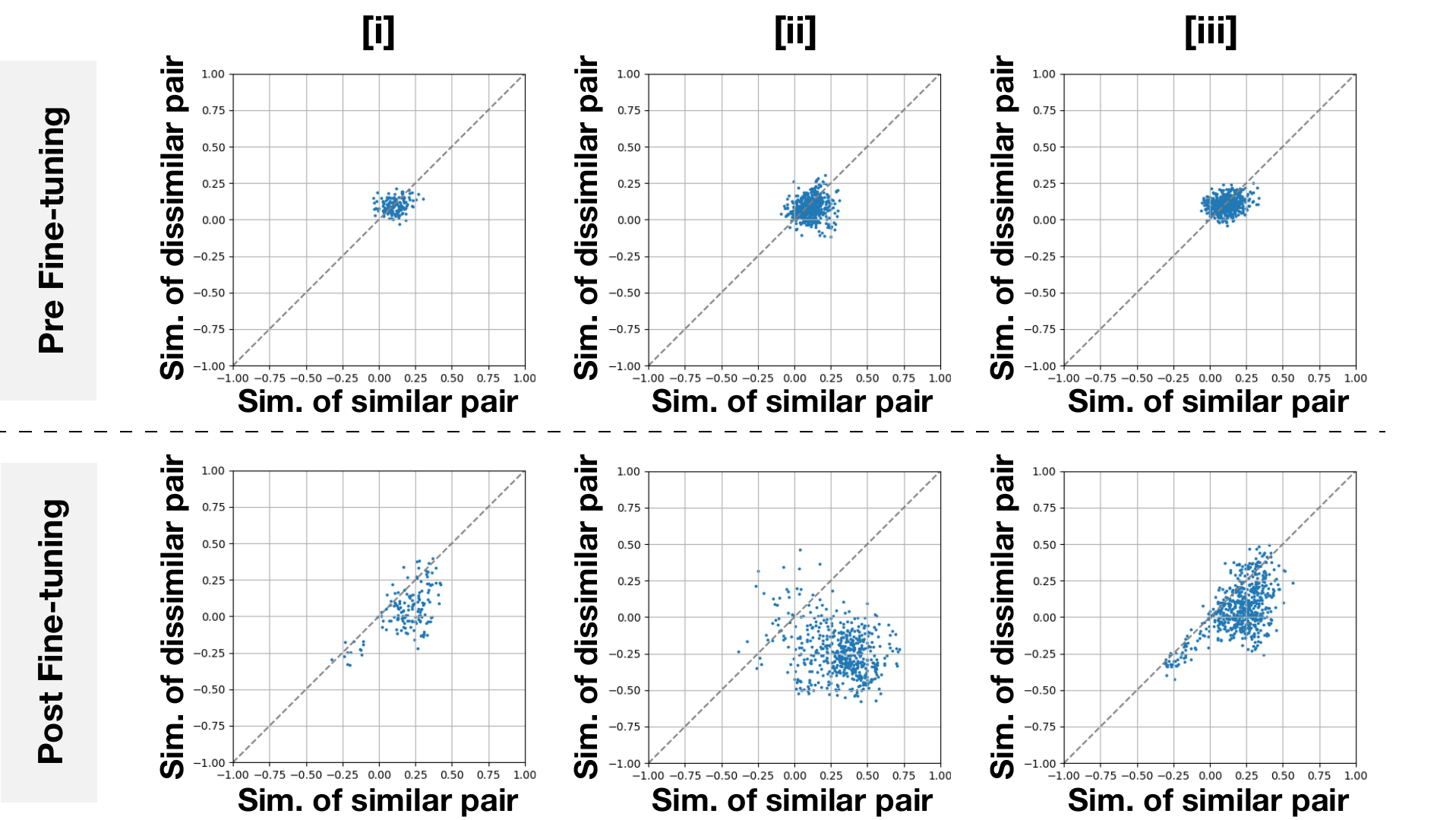}}
\caption{Scatter plot of the similarity scores output by the model for similar and dissimilar pairs in triplet data. }
\label{fig:res-eval}
\end{figure}
Considering the task of face anonymization, there may be cases where the source used for anonymization is known in advance. Therefore, we investigated how prior knowledge of identity affects face similarity evaluation. Additionally, we also examined the impact of training data quality on model performance.

\noindent\textbf{Evaluation Data: [i], [ii], and [iii].}  
We evaluated with three types of evaluation datasets:  
\textbf{[i]} Neither the source nor the target were included in the training data.  
\textbf{[ii]} The target was not included in the training data, but the source was included.  
\textbf{[iii]} The source was not included in the training data, but the target was included.  
As summarized in Fig.~\ref{fig:res-eval}, set [ii] achieved the highest accuracy, suggesting that familiarity with source faces improves performance. In contrast, [i] and [iii], which involved unknown source, resulted in lower accuracy. This highlights the advantage of pretraining on known face-swapping candidates.

\noindent\textbf{Training Data: $D1$ and $D2$.}  
As shown in Table~\ref{tab:acc-10}, we found that while $D1$ benefits from a larger dataset, its inconsistent annotations hinder performance. In contrast, $D2$ achieves higher accuracy due to its superior annotation quality. For the following experiments, $D2$ was employed.

\noindent\textbf{Key Insight.}  
High-quality training data ($D2$) and familiarity with source triplets ([ii]) are critical for optimal model performance, emphasizing the importance of dataset curation and pretraining strategies.

\subsection{Selection of attribute groups}
\label{sec-attr-choose}
This section presents the results of the performance on attribute group classification task.
We determined the most similar attribute group for a given face query image, based on our perceptual similarity metric.

\noindent\textbf{Details of the attribute classification experiment.}
In this experiment, we defined the following eight attribute groups: male, female, young, older, and their combinations: young$\cap$male, older$\cap$male, young$\cap$female, and older$\cap$female. These attribute groups were defined based on annotations from CelebAMask-HQ~\cite{CelebAMask-HQ}. Additionally, we rechecked the annotations and corrected several annotation errors.
For the four subdivided groups (young$\cap$male, older$\cap$male, young$\cap$female, older$\cap$female), 100 face images were randomly selected for each group. For the broader male, female, young, and older groups, each group was constructed as the union of the corresponding subdivided groups, containing 200 face images per group.
By dividing the dataset into these groups, we prepared to analyze how similarity evaluation changes across attribute groups.

For an image \(I_{G_i,j}\) in an attribute group \(G_i\) and a query image \(I_q\), the distance \(d_{I_q,I_{G_i,j}}\) is defined as:
\(
d_{I_q,I_{G_i,j}} = 1 - \cos(f(I_q), f(I_{G_i,j}))
\)
where \( f(\cdot) \) is the feature extractor in our proposed model. The distance between an attribute group \(G_i\) and the query image \(I_q\) is denoted as \(D_{I_q,G_i}\). The attribute group \(G^*\) most similar to the query image \(I_q\) is defined as:
\(
G^* = \underset{G_i}{\arg\min} \, D_{I_q, G_i}
\)

We used a total of 1000 query images, selecting 250 images from each of the attribute groups: young$\cap$male, young$\cap$female, older$\cap$male, and older$\cap$female.
As the evaluation, for each query image $I_q$, the group $G^*$ with the minimum distance was identified, and the accuracy was evaluated by calculating the proportion of queries where the predicted attribute label matched the actual attribute label.

\noindent\textbf{Results.}
\begin{figure}[t]
  \centering
  \centerline{\includegraphics[width=8.5cm]{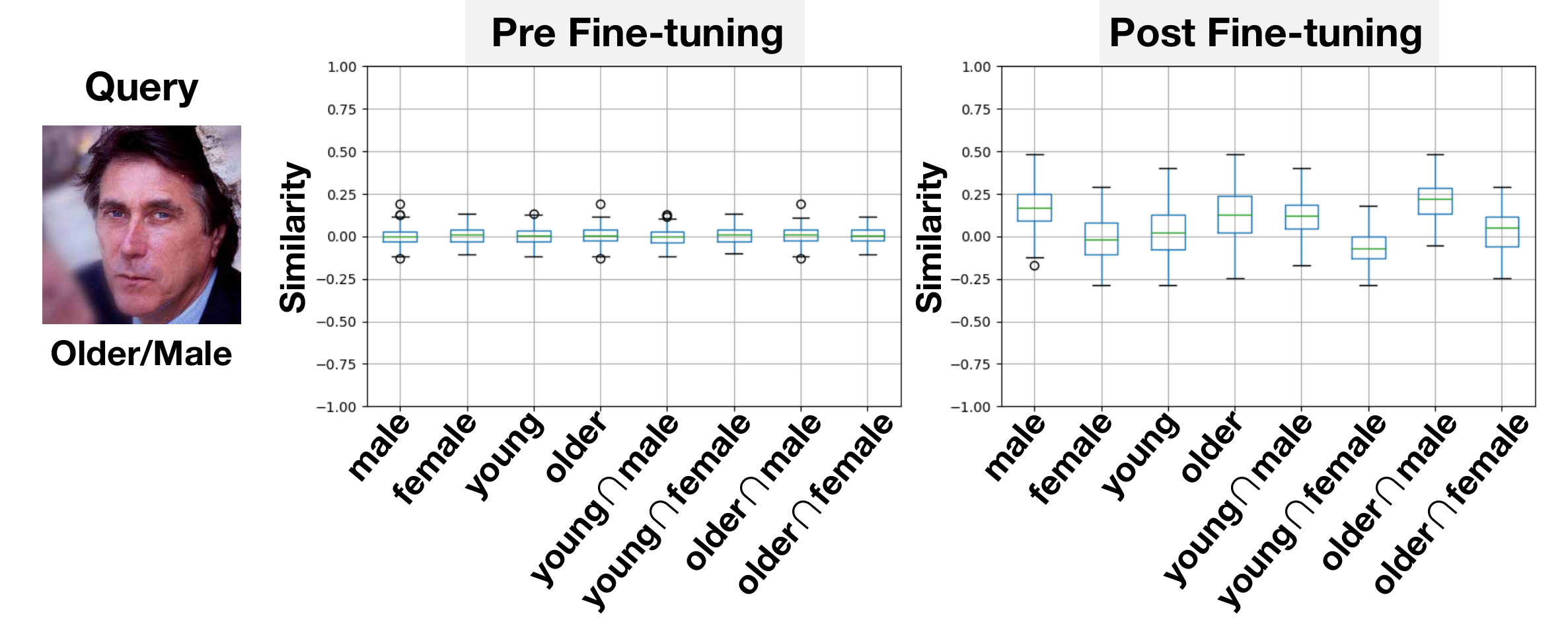}}
\caption{Distributions of similarity between query images and face swap candidates within each attribute group.}
\label{fig:attr-boxplot}
\end{figure}
Fig.~\ref{fig:attr-boxplot} shows the similarity distribution between the query image and each attribute group. Here, we use an image with the attributes “male $\cap$ older” as an example to illustrate the analysis. 
As a result, in the fine-tuned model, the variance within attribute groups and the variance between groups both increased compared to before fine-tuning. It was observed that “male,” “older,” and “male $\cap$ older” had higher similarities than other groups.

Based on the similarity distribution, we classified the closest attribute group to the query image using the 95\% upper confidence interval as the distance. We then verified whether the attribute of the selected group matched the actual attribute of the query image. The classification method was evaluated under two conditions: considering single attribute and considering multiple attributes simultaneously.
For single attribute, we perform binary classification of gender (male or female) and age group (young or older).  
When considering multiple attributes simultaneously, we classify into four groups: “young $\cap$ male,” “young $\cap$ female,” “older $\cap$ male,” and “older $\cap$ female.”

As shown in Table~\ref{tab:attr-II}, after fine-tuning, the AUC increased in most cases compared to before fine-tuning, indicating improved classification accuracy. Focusing on the differences in classification accuracy by attribute, the accuracy for male/female classification is higher than that for young/older classification. 
This suggests that the model prioritizes gender (male/female) over age group (young/older) when evaluating similarity. This can likely be traced back to the annotators' judgments in the training dataset, which may have placed greater emphasis on gender distinctions.

\begin{table}[t]
\footnotesize
\caption{Classification accuracy when the distance \(D_{I_q, G_i}\)is defined as the upper limit of the 95\% confidence interval of distances \( d_{I_q,I_{G_i,j}} \).}
\setlength\tabcolsep{0.3pt} 
\centering

    \begin{tabular}{@{}llcccc@{}}
    \toprule
        \textbf{} & \textbf{Category} & \textbf{Precision} & \textbf{Recall} & \textbf{Accuracy} & \textbf{AUC} \\ \midrule
         \multirow{8}{*}{\shortstack{Pre \\ Fine-tuning}} 
        & Male & 0.930 & 0.936 & \multirow{2}{*}{0.933} & \multirow{2}{*}{0.933} \\ \cline{2-4}
        & Female & 0.936 & 0.930 & & \\ \clineB{2-6}{1.5}
        & Young & 0.829 & 0.812 & \multirow{2}{*}{0.822} & \multirow{2}{*}{0.822} \\ \cline{2-4}
        & Older & 0.816 & 0.832 & & \\ \clineB{2-6}{1.5}
        & Young$\cap$Male & 0.698 & 0.812 & 0.865 & 0.847 \\ \cline{2-6}
        & Young$\cap$Female & 0.807 & 0.652 & 0.874 & 0.800 \\ \cline{2-6}
        & Older$\cap$Male & 0.801 & 0.788 & 0.898 & 0.861 \\ \cline{2-6}
        & Older$\cap$Female & 0.770 & 0.804 & 0.891 & 0.862 \\ \cline{1-6}
        \multirow{8}{*}{\shortstack{Post \\ Fine-tuning}} & Male & 0.959 & 0.972 & \multirow{2}{*}{0.965} & \multirow{2}{*}{0.965} \\ \cline{2-4}
        & Female & 0.972 & 0.958 & & \\ \clineB{2-6}{1.5}
        & Young & 0.871 & 0.754 & \multirow{2}{*}{0.821} & \multirow{2}{*}{0.821} \\ \cline{2-4}
        & Older & 0.783 & 0.888 & & \\ \clineB{2-6}{1.5}
        & Young$\cap$Male & 0.716 & 0.856 & 0.879 & 0.871 \\ \cline{2-6}
        & Young$\cap$Female & 0.822 & 0.664 & 0.880 & 0.808 \\ \cline{2-6}
        & Older$\cap$Male & 0.872 & 0.788 & 0.918 & 0.875 \\ \cline{2-6}
        & Older$\cap$Female & 0.791 & 0.864 & 0.909 & 0.894 \\ \bottomrule
    \end{tabular}

\label{tab:attr-II}
\end{table}


\subsection{Selection of face swap candidates}
\label{sec-face-selection}

\begin{figure}[t]
  \centering
  \centerline{\includegraphics[width=8.5cm]{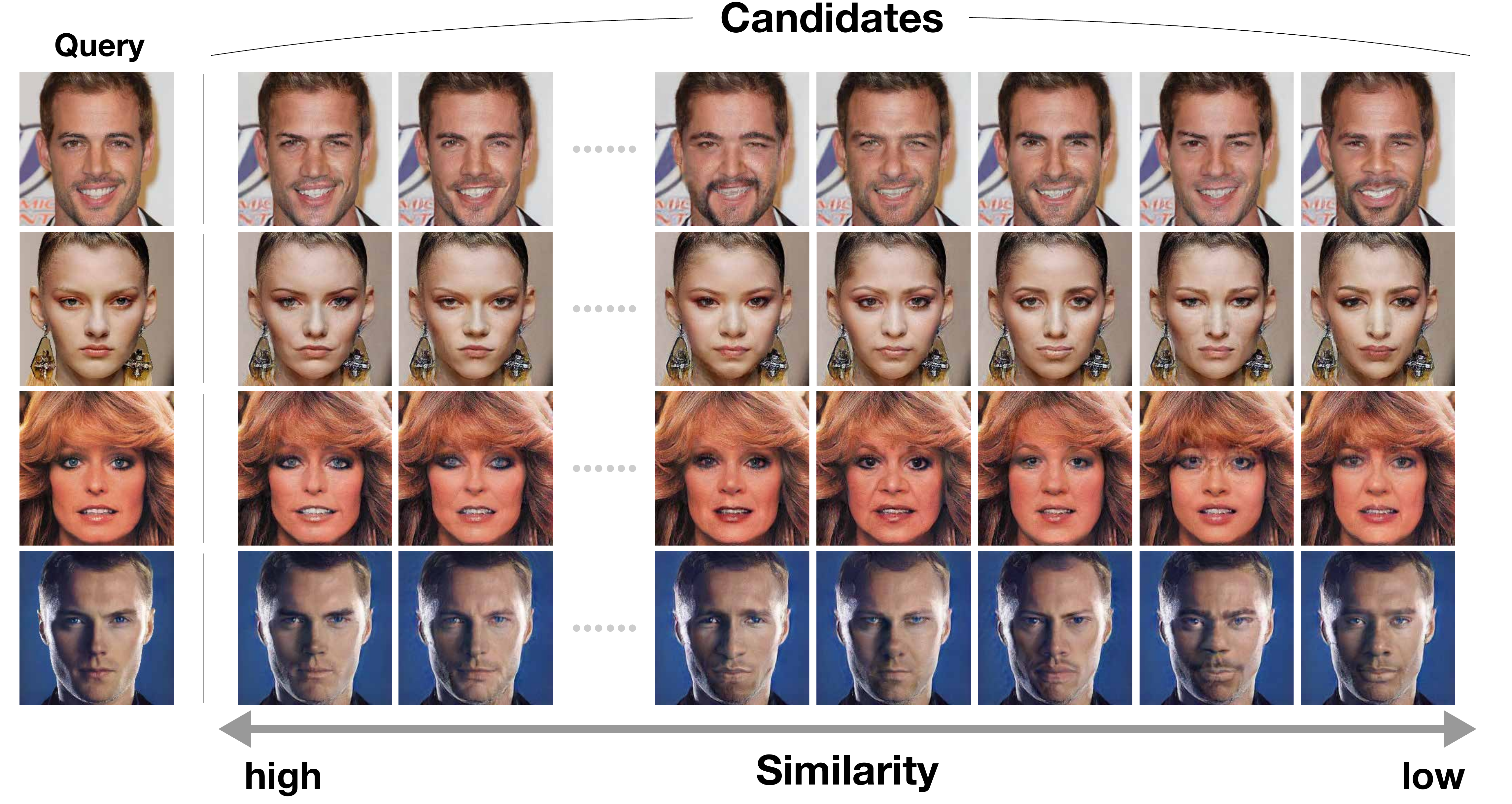}}
\caption{Sorted face swap candidates in the selected attribute group, showing the top two most similar and the bottom five least similar candidates.}
\label{fig:face-selection}
\end{figure}

Based on the attributes of the query image determined in Sec.~\ref{sec-attr-choose}, suitable face swap candidates for anonymization were selected from those sharing the same attributes. To select candidates with lower similarity to the query, the face swap candidates in the selected group were sorted by similarity to the query, as examples shown in Fig.~\ref{fig:face-selection}.

Given the diversity of human faces, even among faces that are not very similar to the query image, various types of “dissimilarity” may exist. To provide users with more flexibility in selecting a “dissimilar” face, this selection algorithm can effectively recommend multiple face swap candidates with relatively low similarity.

\section{Conclusion}

We propose a novel method to address limitations in face similarity prediction for face anonymization via natural face swapping. Conventional methods struggle to evaluate nuanced similarities, such as distinguishing “completely different” from “highly similar but different individuals.”

To overcome this, we introduce a new task to assess “how similar a different identity is,” using our perceptual similarity model PerFace. We constructed an evaluation dataset via user studies and developed a transfer-learning-based model optimized to capture subtle inter-individual similarities.

Our PerFace model significantly outperformed baseline models in face similarity judgment tasks. Additionally, it achieved superior accuracy in attribute classification, highlighting the influence of facial attributes on perceptual similarity judgments and offering insights into perception-based evaluations.

A limitation of this work is its inclusion of subjective factors, especially since facial similarity inherently involves personal perception. Consequently, broader and more diverse perspectives are crucial to capture the sense of similarity that people share. We hope this paper will inspire the creation of larger and more inclusive datasets.


\vfill\pagebreak


\small
\bibliographystyle{IEEEbib}
\bibliography{strings,refs}

\end{document}